%% file: paper.tex
\documentclass{llncs}
\usepackage{epsf}

\usepackage{multibib}
\newcites{languageresource}{Language Resources}
\usepackage{graphicx}
\usepackage{tabularx}
\usepackage{soul}
\graphicspath{ {images/} }
\usepackage{epstopdf}
\usepackage[latin1]{inputenc}

\usepackage[T1]{fontenc}
\usepackage{stackrel}
\usepackage{amsmath,amssymb}
\usepackage{hyperref}
\usepackage{xstring}
\usepackage{latexsym}
\usepackage{url}

\begin{document}

\title{OntoSenseNet: A Verb-Centric Ontological Resource for Indian Languages} 

\author{
Jyoti Jha\thanks{author has contributed to resource development for Hindi}, Sreekavitha Parupalli\thanks{author has contributed to resource development for Telugu}, Navjyoti Singh}
\institute{
           Center for Exact Humanities\\
           IIIT-Hyderabad\\
           jyoti.jha@research.iiit.ac.in, sreekavitha.parupalli@research.iiit.ac.in, navjyoti@iiit.ac.in}

\maketitle

\begin{abstract}
Following approaches for understanding lexical meaning developed by Y{\=a}ska, Patanjali and 
Bhartrihari from Indian linguistic traditions and 
extending approaches developed by Leibniz and Brentano in the modern times, a framework of formal ontology of language was developed.
This framework proposes that meaning of words are \textit{in-formed} by intrinsic and extrinsic ontological structures. The paper aims to capture such intrinsic and extrinsic meanings of
words for two major Indian languages, namely, Hindi and Telugu. Parts-of-speech have been 
rendered into sense-types and sense-classes. Using them we have developed a 
gold-standard annotated lexical resource to support semantic understanding of a language. 
The resource has collection of Hindi and Telugu lexicons, which has been manually 
annotated by native speakers of the languages following our annotation guidelines. 
Further, the resource was utilised to derive adverbial sense-class distribution of verbs and k{\=a}raka-verb sense-type distribution.
Different corpora (news, novels) were compared using verb sense-types distribution. Word Embedding was used as an aid for the enrichment of the resource.
This is a work in progress that aims at lexical coverage of 
language extensively.\\

\end{abstract}

\section{Introduction}
The concept of `meaning' has been discussed for a long time. Cognitively it can be understood to have an
\textit{intensional} or \textit{extensional} form. Frege~\cite{gamut1991logic} discussed the idea of sense
and reference. He called `sense' as \textit{intensional} meaning and `reference' as \textit{extensional}
meaning. The meaning that has a constant value in an
expression is \textit{intensional}, whereas the meaning that is contributed  by the real world to the
mental concept is \textit{extensional}. Two words are said to be extensionally equvivalent if they refer to the
same set of objects, whereas if they share the same features then they are intensionally equivalent. According
to Frege every signficant linguistic expression has both 'sense' and 'reference'. The other theories of meaning
are correspondence theory, consensus theory, constructivist theory etc. 
All these account for extensional meaning.
\par Meaning of a word in a language is generally derived from dictionary or from a context it is used in.
Speaking from an ontological viewpoint, the meaning of a word can be understood based on its participation in classes, events and relations.
In order to manipulate language computationally at the level of lexical meanings,
Otra~\cite{otra2015towards} developed Formal Ontology of Language.
It considers meaning to have an intrinsic form. According to the theory proposed, meanings have primitive ontological forms.
It is language independent and aims at extensive coverage of language.
\par This paper uses the idea of Formal Ontology of Language to develop lexical resource for Hindi
and Telugu. 
Section 2 discusses the previous works that have been done in order to specify meaning of a word and
development of various lexical resource.
Section 3 of the paper discusses the Formal Ontology of Language, as proposed by Otra~\cite{otra2015towards}.
Section 4 talks about data acquisition for Hindi and Telugu and it shows how sense identification is done for different parts-of-speech.
k{\=a}raka information for sense-types of verbs have been extracted from Hindi corpus. OntoSenseNet,
a user interface has been built for our ontological resource.
Section 5 shows validation of the resource based on inter-coder agreement.
Section 6 demonstrates enrichment of the resource using word embeddings.
Representation of verbs through their adverbial class distribution has been studied.
Different corpora(news, novels) have been compared using frequency profiling of verb sense-types.
Section 7 outlines conclusion and future work. 
IAST based transliteration\footnote{\url{https://en.wikipedia.org/wiki/International\_Alphabet\_of\_Sanskrit\_Transliteration}} for Devanagari and Telugu scripts has been used in the paper.
\section{Previous Work}
In this section we discuss several attempts that have been made by various researchers in order to specify
meaning of a word. Considering verb as the core of a language, some linguists derived different classifications.
Along with different kinds of verb classifications, there have been approaches to derive semantic primitives.

\par{\textbf{Levin's classification}}
Levin assumed that syntactic behavior of a verb is semantically determined~\cite{levin1993english}.
He classified meanings of about 3,000 English verbs. They were composed into 50 primary-classes and 192
sub-classes using methodical study of 79 diathesis alternations. It mainly deals with verb taking noun
and prepositional phrase complements. Although it can empirically classify verbs but it only captures some facets
of semantics~\cite{korhonen2004extended}.
It does not include verbs taking ADJP, ADVP, predicative, control and sentential complements and 
is highly language dependent. 

\par{\textbf{VerbNet classfication}}
VerbNet ~\cite{kipper2000class} is a hierarchical verb lexicon that represents verbs syntactic
and semantic information. In each verb class, thematic roles are used to link syntactic
alternations to semantic predicates. However, it contains limited coverage of lemmas and for each lemma the coverage of the senses are limited.
Since  it has been inspired from Levin's verb classification it is also language dependent.

\par{\textbf{WordNet}}
It is a lexical database inspired by psycholinguistic theory of human lexical memory ~\cite{miller1990introduction}.
Words are organized as synsets. These synsets represent lexicalised concepts which are organized into synonym
sets. These synsets are connected to other synsets by means of semantic relations.
Nouns are organized as hypernymy and hyponymy relations. Verbs are organized as hypernym, troponym,
entailment and coordinate terms. It does not have classification of adverbs.
It also lacks information about verb syntax and is also language specific.

\par{\textbf{Wierzbicka's semantic primitives}}
The concepts that can be innately understood without any further decomposition are Semantic Primes. 
The widely used example to explicate this concept is the verb `\textit{touching}'. Its meaning can be readily understood, 
however a dictionary might define `\textit{touch}' as "to make contact" and `\textit{contact}' as "touching",
provides no information if neither of these words are understood. The theory of semantic primes was introduced
by Wierzbicka, ~\cite{wierzbicka1972semantic}. It has been criticised for its reductive approach and is
limited by its generative coverage in any language ~\cite{riemer2006reductive}.
\par The limitations of the above theories in terms of being language specific and limited coverage in a language
led to the formulation of Formal Ontology of Language by Otra ~\cite{otra2015towards}. In the proposed
theory, the meaning are considered to have primitive ontological forms and they are independent of a language.
Each of the parts-of-speech are organised as \textit{types} or \textit{classes}. To derive the intensional
meaning of a sentence, one has to consider the relation between different parts-of-speech e.g the relation
between verb-adverb, noun-adjective, verb-noun. The relations between the points are also considered to have
ontological forms, which can help specifying meaning at a sentence level.
The next section discusses the theory of Formal Ontology of Language, as introduced by Otra.
\section{Formal Ontology of Language}
In a language one can describe a state of affairs using different verbs, hence there is a verbal ambiguity. To derive the universal verb there have been several discussions in Greek and Indic traditions.
While in Greek tradition `be' is considered as the universal verb ~\cite{kahn1973verb}, on the other hand
Indic tradition considers `happening (bhavati)' as the universal verb.
Let us consider a question "what are you doing?". This can be answered with the verb `do'. Hence one can say
that `do' can be a primitive sense that will be present in every verb. However, Patanjali mentions three verbs
that cannot be the answer to the above question. These are (1) being/existence (asti), (2) presence (vidyate), (3)happening
(bha{\=a}va). According to Bhartrihari, verb has sense of sequence and state. Hence, it has a sense of happening making it the universal verb. 
Linguistic traditions in India have long regarded verb as the centre of language (in both 
syntactic and semantic terms) right from Y{\=a}ska , 
P{\=a}nini, Patanjali and 
Bhartrihari  ~\cite{yudhishtir}, 
~\cite{staal1972reader}, ~\cite{potter2014encyclopedia}.
Meaning of verbal element is seen as \textit{bh$\overline{a}$va} (happening) as opposed to 
\textit{satt$\overline{a}$} (being) which stands behind nominal elements 
~\cite{bhattacharya1958yaska}, ~\cite{ogawa2005process}. Later, independent of 
linguistic discourse, logicians brought out hundreds of \textit{bh$\overline{a}$va-s} 
(happening) with atomic transformational structure 
\textless $entity_1$|$entity_2$\textgreater   like cause|effect, part|whole, predecessor|successor, qualifier|qualified, 
ascribed|ascriber, locus|located, etc. These are atomic discriminants which form 
elementary meanings. 
Meanings are not seen as an entity like semantic primitive ~\cite{wierzbicka1972semantic} 
but as a unified discriminative structure with a form <$entity_1$|\textit{contiguous with}|$entity_2$, in 
context of \textit{continuum}>. For example, verb `move' has a sense <predecessor state|\textit{contiguous with}|successor state, in context of `move' \textit{continuum}>. Its meaning is
a continuant feel of motion punctuated by discriminating logical structure of predecessor and successor states. 
One can read in its meaning such discrimination points. Leibniz called such punctuations as 
actual points ~\cite{leibniz2001labyrinth} as endeavours, as ontologically vacuous, as 
different from Euclidean points. Brentano ~\cite{brentano2009philosophical} also built an
idea of mental continuants as punctuated with \textit{modo recto} and \textit{modo obliquo}. 
These boundaries, punctuations or points are ontological as they vacuously discriminate ontic
entities or states which are felt as continuous. Otra ~\cite{otra2015towards}  built formal ontology of lexical meaning using such punctuational boundaries.

Meaning of `move' is always more than the discriminative senses we read in it.  In the proposed formal ontology
seven discriminant punctuations that are read in all verbs ~\cite{otra2015towards} are suggested and
determined.
When we say, `rapidly move' or 'hesitantly move', we have done adverbial modification of the 
meaning of `move' and have added new modifier|modified points in its meaning. When appending
`rapidly' we add temporal-feature-class in adverb whereas while appending `hesitantly' we add force-feature-class in adverb to the meaning of `move'.
Even when we have discriminated temporal or force features, meanings of `rapidly' and `hesitantly' are more
than their adverb sense-classes. Otra ~\cite{otra2015towards} has delineated four adverb classes of 
discriminant point.
Verbs are also seen as contiguants of nouns in seven or eight case relations.
These seven/eight classes of verb-noun pairing are further coincident boundaries in the meaning of articulation
with the verb.
Further, noun-noun pairs and noun-adjective pairs are more coincident 
points. Otra ~\cite{otra2015towards} also proposes twelve noun-verb types of sense-points in the ontology.
The verb-centric formal ontology of meaning is based on sense-type and sense-class of 
punctuational boundaries that can be located in lexical meaning. TYPES and CLASSES are logical forms of intensional
senses [~\cite{otra2015towards}, page 13,14,15].
Using the formal ontology we are building
lexical resource of verbs, adverbs, nouns and adjectives in terms of basic discriminant points, which
\textit{in-form} their meaning. The formal ontology is language independent
and thus we are developing the resource for several languages at once.
\par In the next section we discuss the resource creation for Indian languages, namely Hindi and Telugu using the
proposed Formal Ontology of Language. 
\section{Resource Building}
Otra ~\cite{otra2015towards} has developed the resource for English that has 3,867 verbs, 1,980 adverbs and 300 adjectives. 
In our resource, Sense-types of 3,152 Hindi and 3,379 Telugu verbs has been manually identified. 
Similarly manual identification of sense-classes of 2,214  Hindi and 101 Telugu adverbs has been done.
Sense-types of 238 Hindi adjectives has been identified.
Annotation for sense-types of Telugu adjectives is in progress.
The annotators have native proficiencies in the corresponding languages. 
\subsection{Data Acquisition}
Words were collected from different resources like dictionary, wordnet. These were further 
used to populate our resource. Since this is a work in progress, not all the words from 
the different resources have been added into our resource.  
\par{\textbf{Hindi}} 
Distinct verbs, adverbs and adjectives for Hindi were collected from Hindi Wordnet 
\footnote{\url{http://www.cfilt.iitb.ac.in/wordnet/webhwn/}} and Dictionary 
\footnote{\url{https://ia601603.us.archive.org/20/items/in.ernet.dli.2015.348711/2015.348711.Hindi-Shabdasagar.pdf}}. 
\par{\textbf{Telugu}}
Telugu being a resource poor language, does not have a usable soft copy of Telugu-Telugu dictionary till date.
We are developing it from the printed copy of "Sri Suryaraayandhra Telugu Nighantuvu" ~\cite{pantulu1936sri}
for all of its eight volumes. 
Verbs, adverbs and adjectives have been completely populated in the usable soft copy from this dictionary, whereas work is still under progress for
other parts-of-speech.
Dictionaries for Telugu-Hindi, English-Telugu are available\footnote{\url{https://ltrc.iiit.ac.in/onlineServices/Dictionaries/Dict\_Frame.html}}.

Table~\ref{tab:table1} shows the number of distinct  verbs, adverbs and adjectives in each of the resources.

\begin{table}
\caption{Distinct number of verbs, adverbs and adjectives for Hindi, Telugu in different resources}
\label{tab:table1}
\begin{center}
\begin{tabular}{rrrr}
\hline
\bf Resource & \bf Distinct Verbs & \bf Distinct Adverbs & \bf Distinct Adjectives\\
\hline
Hindi Wordnet & 6778    & 2114  &  19190\\
Hindi-Shadsagara Dictionary & 3529 & 1650 & 36398  \\
OntoSenseNet(Hindi) & 3152 & 2214 & 238 \\
Telugu-Telugu Dictionary & 8483 & 253&11305\\
Telugu Wordnet\footnotemark & 2795  & 442& 5776\\
Telugu-Hindi & 9939 & 142& 1253\\
OntoSenseNet(Telugu) & 3379 & 101&  \textit{Annotations are yet to be started}\\
\hline
\end{tabular}
\end{center}
\end{table}
\footnotetext{\url{http://tdil-dc.in/indowordnet/}}

\subsection{Sense-Identification}
\par{\textbf{Verbs}}
Different verbs can be used for describing the same situation. Thus verbs are colocative in nature.
In a single verb many verbal sense points can be present and different verbs may share same verbal
sense points. For example "walking", "running" entails a sense of `move'. Verb like "studying"
entails a sense of `know' and `do'. "Eating" entails a sense of `do' and `have'.
Thus, verbs are organised as sense-types. Otra ~\cite{otra2015towards} has shown the existence of seven primitive sense-types
of verbs. These seven sense-types of verbs have been derived by collecting the fundamental 
verbs used to define other verbs. These verbs were then grouped using intrinsic senses, 
which were designated to a particular sense-type. These sense-types are inspired from 
different schools of Indian philosophies. The seven sense-types of verbs are listed below 
with their primitive sense along with two Hindi and Telugu examples each.
\begin{enumerate}
\setlength\itemsep{0em}
\item Means|End - Do; khelan{\=a} (play), karan{\=a} (do); {\=a}duta (play), ceyuta (do) 
\item Before|After - Move; bahan{\=a} (flow), calan{\=a} (walk); p{\=a}ruta (flow), naduvuta (walk)
\item Know|Known - Know; j{\=a}nan{\=a} (know), parakhan{\=a} (examine); {\=u}himcuta (imagine), paris{\=i}limcuta (examine)
\item Locus|Located - Is; rahan{\=a} (stay), hon{\=a} (happen); umduta (to be, stay), jaruguta (happen) 
\item Part|Whole - Cut; k{\=a}tan{\=a} (cut), mit{\=a}n{\=a} (erase); koyuta (cut), vidipovuta (separate)  
\item Wrap|Wrapped - Cover; jh{\=a}mpan{\=a} (cover), pahan{\=a}n{\=a} (dress-up someone); m{\=u}yuta (cover) , {\=a}kramimcuta (contain forcefully) 
\item Grip|Grasp - Have; p{\=a}n{\=a} (get), len{\=a} (take); bhayapaduta (fear), t{\=i}sukonu (take)
\end{enumerate} 
Each of the verbs can have all the seven dimensions of sense-types. The degree depends on the
usage/popularity of a sense in a language. In our resource we have identified two sense-types
of each verb, i.e. primary and secondary.
Consider the verb `dance' in the sentence "Madhuri is dancing gracefully". 
Here `dance' involves a sense of movement which a doer does. Thus Before|After is a primary 
sense and Means|End is a secondary sense.
For polysemous verbs, sense-type identification was done for each of their different meanings.
For example, the verb "rap" has three meanings. Thus rap1, rap2, rap3 have been added in the 
resource along with its meaning sense-types.
\begin{enumerate}
\item rap1- Criticizing someone, Means|End and Know|Known.
\item rap2- To perform rap music, Means|End and Before|After.
\item rap3- To hit or say something suddenly and forcefully, Means|End and Part|Whole. 
\end{enumerate}
Sense-types for 3,152 Hindi and 3,379 Telugu verbs were manually identified.
Figure 1 shows the sense-type distribution for English, Hindi and Telugu verbs in OntoSenseNet.
\begin{figure}[h]
\includegraphics[width=8cm,height=30cm,keepaspectratio] {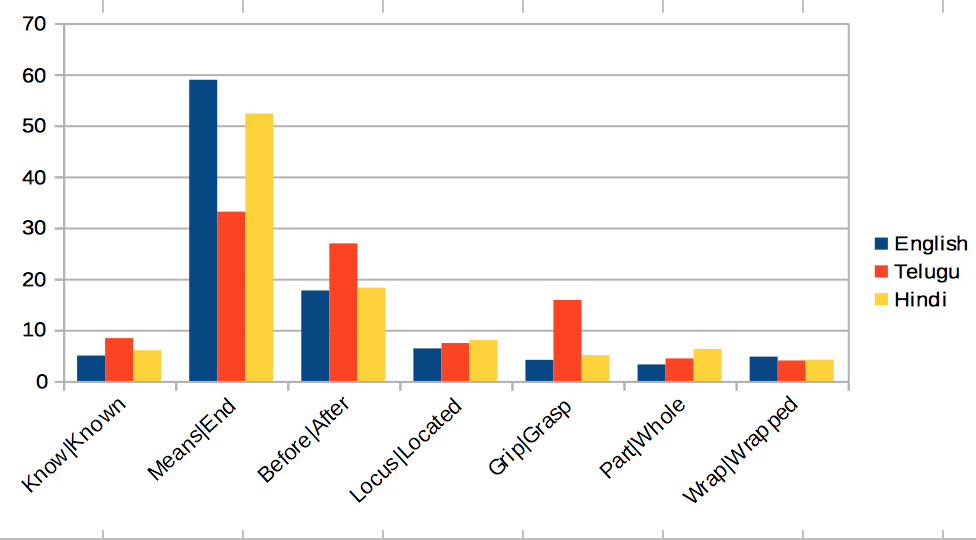}
\caption{ Verb Sense-type distribution}
\end{figure}
\par{\textbf{Adverbs}}
Meaning of verbs can further be understood by adverbs, as they modify verbs. The sense-classes of adverbs
are inspired from adverb classification in Sanskrit. Following are the identified sense-classes along with their fundamental sense, illustrated with English, Hindi and Telugu examples 
\begin{enumerate}
\setlength\itemsep{0em}
\item Temporal - Adverbs that attributes to sense of time. e.g Never; sasamaya (timely); varusag{\=a} (continuously) 
\item Spatial - Adverbs that attributes to physical space. e.g There; p{\=a}s (near); davvu (far away)
\item Force - Adverbs that attributes to cause of happening e.g. Dearly; barbas (unwillingly); gattiga (tightly)
\item Measure - Adverbs dealing with comparison, judgement. e.g - Only,; lagbhag (approximately); gaddu (abundantly)
\end{enumerate}
Sense-classes for 2,214 Hindi and 101 Telugu adverbs have been manually identified.
Table~\ref{tab:table2} shows sense-class distribution of adverbs for English, Hindi and Telugu.
\begin{table}
\caption{Adverb Sense-Class Distribution}
\label{tab:table2}
\begin{center}
\begin{tabular}{rrrr}
\hline
\bf Sense-Class&\bf English & \bf Hindi & \bf Telugu \\
\hline

Temporal & 5.5& 24.3& 28.7\\
Spatial & 2.7 & 13.5& 12.8  \\
Measure & 39.4 &32.2& 31.6 \\
Force & 52.2  &30 &  26.7\\
\hline
\end{tabular}
\end{center}
\end{table}

Sub-classfication of adverb sense-classes is being developed. 
\par{\textbf{K{\=a}raka Relation}}
K{\=a}raka are classes that are used for expressing relation between words in a 
sentence. Computational Paninian Grammar Framework describes k{\=a}rakas as 
syntactico-semantic (or semantico-syntactic) relations between the verbs and their 
related constituents (generally nouns) in a sentence ~\cite{bharati2007computational}.
It describes eight types of k{\=a}rakas:-
k1: kart{\=a} (Nominative), k2: karma (Instrument), k3: karna (Ablative), k4:samprad{\=a}na (Possessive), k5:ap{\=a}d{\=a}n (Objective), k6:sambandh (Dative), k7:adhikaran (Locative), k8:sambodhan (Vocative).
Distribution of verb sense-types and k{\=a}raka were studied in  Hindi novel corpora containing 3,39,057 words.
This corpus was collected from Hindisamay\footnote{\url{http://www.hindisamay.com/}} and was fully parsed using ISCNLP tagger\footnote{\url{https://github.com/iscnlp/isc nlp}}.
For Telugu, development of treebank data and full dependency parser is still under process. Thus, K{\=a}raka-Verb sense-types distribution of Telugu has not been extracted.
Figure 2, shows verb sense-types distribution for different k{\=a}raka in Hindi. It shows that Locus|Located type
of verbs have mostly occurred with a k1 relation with noun, whereas the k4 k{\=a}raka is hardly occurs in any verb-noun relation.
\begin{figure}[h]
\includegraphics[width=10cm,height=15cm,keepaspectratio] {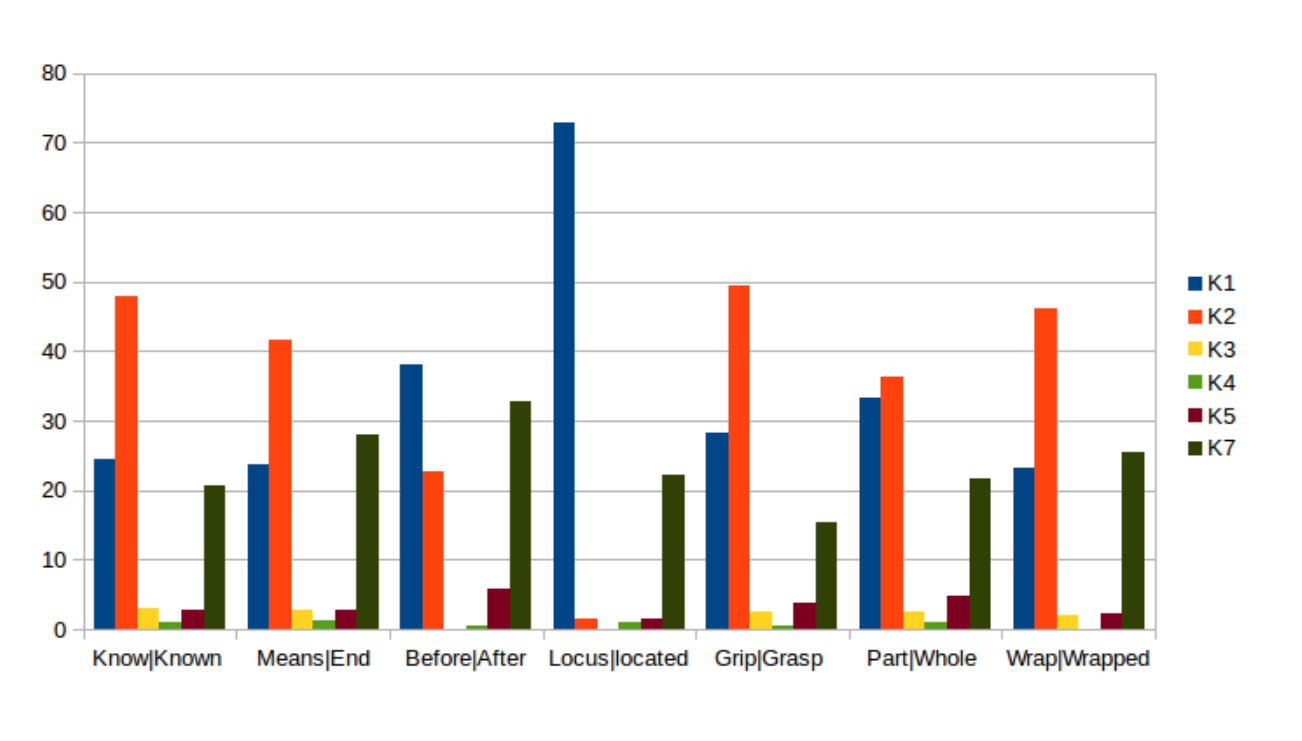}
\caption{ Verb sense-type and Karaka Distribution}
\end{figure}
\par{\textbf{Adjectives}}
Like verbs, adjectives are also colocative in nature. Otra ~\cite{otra2015towards} 
identifies 12 sense-types. However these can be reduced to 6 pairs. Following are the identified six sense-types pairs of adjectives along
with their meanings and one English, Hindi  and Telugu examples each. 
\begin{itemize}
\item Locational - Adjectives that universalise or localise a noun e.g Local, doosr{\=a} (Other), nirdista (specific)
\item Quantity -  Adjectives that either qualify cardinal measure or quantify in ordinal-type  e.g. - only, eka (one), okkati (One) 
\item Relational -  Adjectives that qualify nouns in terms of dependence or dispersal e.g. similar, m{\=a}trih{\=i}n  (without mother),
vistrta (broad) 
\item Stress - Adjectives that intensify or emphasis a noun - e.g, strong, mazb{\=u}t (strong), gatti (strong)
\item Judgement - Adjectives that qualify evaluation or qualify valuation feature of a noun e.g - bad, acch{\=a} (good), mamci (good)
\item Property - Adjectives that attribute a nature or qualitative domain of a noun. e.g - black, k{\=a}l{\=a} (black), nallani (black) 
\end{itemize}
Sense-types for 238 Hindi adjectives have been manually annotated. Sense-types for Telugu is yet to be started. 
\subsection{User Interface}
A user interface, OntoSenseNet 
has been developed for this ontological resource. Input of a verb/adverb/adjective returns 
its corresponding sense-types/classes along with its illustrative meaning and example sentences.
Further, development on crowdsourcing of the resource is being done where users can login and populate the data
by providing examples to support their claims. This will be manually verified before adding to OntoSenseNet.
For divided opinion, a discussion page would be provided.
\section{Resource Validation}
To show the reliability of the resource, Cohen's Kappa
~\cite{carletta1996assessing} was used to measure inter coder agreement. The annotation was done
by one human expert and it was cross-checked by another annotator who was equally trained.
Verbs and adverbs were randomly selected from our resource for the evaluation
sample.
The inter coder agreement for 500 Hindi verbs and 1,000 adverbs were 0.70 and 0.91 respectively.
Similarly validation for 500 Telugu verbs was done, for which inter coder agreement was 0.82.
Validation for both the language resources shows high agreement.
~\cite{landis1977measurement:22}. Further validation of the resource is in progress.
\section{Resource Enrichment and Utilization}
\subsection{Sense-Identification of Verbs and Adverbs using Word Embeddings}
Word Embeddings have been widely used for extracting similar words 
~\cite{leeuwenberg2016minimally:29}. Previous study has shown that word embedding
has significant improvement over WordNet based measures ~\cite{singh2014merging:17}. 
We used this property to assign  sense-type of verbs and sense-class of adverbs. This was done in order to
facilitate the annotation task. However, this was further verified
manually.

\par{\textbf{Method} Hindi corpus was collected from Leipzig 
\footnote{\url{ http://corpora2.informatik.uni-leipzig.de/download.html}}, Hindi wiki-dump
\footnote{\url{http://kperisetla.blogspot.in/2013/01/wikipediahi-offline-wikipedia-in-hindi.html}} and Lindat ~\cite{11372/LRT-987:19}.
It contains 3,73,45,049 sentences and 75,31,64,082 words.
This corpus was fully parsed using iscnlp tagger\footnote{\url{https://github.com/iscnlp/iscnlp}}.
Word2vec ~\cite{mikolov2013distributed:18} was trained on this corpus using CBOW technique and vector dimensions were 100.1,485 verbs and  1,054 adverbs were randomly chosen from the corpus for the sense-identification. Out of these, sense-type of
1,182 verbs  and sense-class of 832 adverbs were already present in the resource. The sense identification for the remaining words
were carried out.In order to identify sense of a word, its cosine similarity was calculated against the words whose sense were already present
in the resource. Cosine similarity above 0.7 was considered. The maximum occurring sense in the similarity cluster was considered
to be the potential sense of that word. This was subsequently verified manually.
For example, sense-type of the Hindi verb 'c{\=i}ran{\=a}' (tear) was not present in
the resource(OntoSenseNet). The sense-type of those verbs were considered whose cosine similarity with 'c{\=i}ran{\=a}' was above 0.7.  
The maximum occurring sense from these set of verbs was Part|Whole. Thus, the sense-type of 'c{\=i}ran{\=a}' was assigned as Part|Whole. The above method was executed to identify sense-type of 303 verbs and sense-class of 222 adverbs. This method correctly identfied the sense
-class of 220 adverbs and sense-type of 185 verbs. The sense identified for the words were finally incorporated in the resource.
Table~\ref{tab:table3} summarises the statistics. Column A is part-of-speech. Column B shows number of words in that part-of-speech that were
randomly sampled from the corpus. Column C shows number of words for which sense were already present in the resource. Column D shows number of words for which sense identification was carried out. Column E contains number of words whose sense
were correctly identified by Word2Vec. Column F shows accuracy in percentage.

\begin{table}
\caption{Statistics for the sense-identification by Word2Vec}
\label{tab:table3}
\begin{center}
\begin{tabular}{rrrrrr}
\hline
\bf A & \bf B& \bf C& \bf D& \bf E& \bf F\\
\hline
Verb & 1,485 & 1,182 & 303 & 185 & 61.056\%\\
Adverb & 1,054 & 832 & 222 & 220 & 99.09\%\\
\hline
\end{tabular}
\end{center}
\end{table}

Table ~\ref{tab:table4} and Table ~\ref{tab:table5} shows the verb and adverb clusters, respectively. In each of the tables the similarity with the words in column-1 is 
above 0.7. Column 3 shows the maximum occurring sense-type/sense-class 

\begin{table}
\caption{Similarity Cluster and the maximum occurring sense-type}
\label{tab:table4}
\begin{center}
\begin{tabular}{rrr}
\hline
\bf Verb & \bf Verb-Clusters & \bf Maximum occurring sense-type\\
\hline
\textbf{c{\=i}ran{\=a}} & nocan{\=a}, ghisan{\=a}, chedan{\=a}, khuracan{\=a}, p{\=i}san{\=a},phul{\=a}n{\=a} & Part|Whole\\
\textbf{j{\=a}nan{\=a}} & bat{\=a}n{\=a}, kahan{\=a}, len{\=a}-den{\=a}, m{\=a}l{\=u}ma, m{\=a}nan{\=a}, p{\=u}chan{\=a}  & Know|Known\\
\hline
\end{tabular}
\end{center}
\end{table}

\begin{table}
\caption{Similarity Cluster and the maximum occurring sense-class}
\label{tab:table5}
\begin{center}
\begin{tabular}{rrr}
\hline
\bf Adverb & \bf Adverb-Clusters & \bf Maximum occurring sense-class\\
\hline
tigun{\=a}& dogun{\=a},dugun{\=a},caugun{\=a}& Measure\\ 
yak{\=a}yaka& sahas{\=a},ek{\=a}eka,ac{\=a}naka &Temporal\\
\hline
\end{tabular}
\end{center}
\end{table}

\subsection{Representation of verbs through adverbial semantics}
Representation of verbs as a combination of their participatory adverb modifiers has not been exhaustively
studied till now. Using our resource one can study the adverbial features of verb. 
We extracted Verb-Adverb relation from a fully parsed corpora.
\par Using full dependency parser of Hindi, 25,00,130 sentences were parsed. Verbs whose frequency was above 50
were considered in order to extract their modifying adverbs. 
The sense-class of these adverbs were then identified using our resource. Percentage of the frequency
distribution of these sense-classes of adverbs for every verb was calculated. 
Table ~\ref{tab:table6} shows few examples of
represenation of verbs in terms of their frequency distribution of adverb sense-class.

\begin{table}
\caption{Percentage of frequency distribution of adverb sense-class of verbs}
\label{tab:table6}
\begin{center}
\begin{tabular}{rrrrr}
\hline
\bf Verb& \bf Temporal& \bf Measure&\bf Spatial&\bf Force\\
\hline
cunan{\=a} & 60.82&36.08&2.06&1.03\\
j{\=a}n{\=a} &61.12&25.92&12.96&0\\
calan{\=a}& 44.26 &44.26&6.55&4.91\\
likhan{\=a}&32.07 &50.31&10.69&6.91\\
\hline
\end{tabular}
\end{center}
\end{table}
Few examples of the verbs that
were not modified by "Spatial" in the corpora are karw{\=a}ne [to make someone do], chaunk [to be surprised], 
bach{\=a} [save], gher [circle]
It is interesting to note that spatial and force  were the only classes that did not 
modify some verbs. 
OntoSenseNet has verb-adverb pairing frequencies also.
\par {\textbf{Corpora Comparison}} Frequency distribution of verb sense-type and adverb sense-class across different types of corpora(novel, news) have been statistically studied. Previous works have shown the usage 
of frequency profiling ~\cite{rayson2000comparing:20} for comparing different corporas.
We have used this approach to identify key ontological points that differ across corpora.
Log-Likelihood estimation was calculated using contingency table for each of the verb sense-type.
It was observed that Means|End sense-type is the most indicative of the news corpora
that was collected using Hindi Treebank(3,65,431 words).

\begin{table}
\caption{Frequency distribution and Log Likelihood}
\label{tab:table7}
\begin{center}
\begin{tabular}{rrrrr}
\hline
\bf Sense-Type & \bf Novel & \bf News & Log-Likelihood \\
\hline
Means|End    &  25.215  &  38.270  & +38523.04\\
Before|After  &  19.084  &  15.293   & +9787.00   \\
Part|Whole  & 5.917   &  5.736  & +4290.64   \\
Grip|Grasp  & 7.387  & 9.782  & +9076.68   \\
Locus|Locate  & 30.817  & 23.946 & +14911.13\\
Know|Known  & 10.216 & 5.993 & +2812.73 \\
Wrap|Wrapped  & 1.360  & 0.977 & +566.23\\
\hline
\end{tabular}
\end{center}
\end{table}
Table ~\ref{tab:table7}  shows the frequency distribution of sense-types of verbs and their log-likelihood.
\par Furthermore, adverbial class distribution was observed in novels of two different authors.
The adverbial distribution was considered for the most commonly occurring verbs in
both the corpora. The difference in the use of adverbs may be accounted for different sociolinguistics aspects and can be applied in the study of social differentiation in the use of a language.
Adjectival and k{\=a}raka information needs to be exploited further for a deeper insight into corpora comparison.Sense-identification for Telugu using Word2vec is in progress.
\
Table ~\ref{tab:table8} shows few examples of the adverbial sense-class distribution
for the verbs used by both the authors.

\begin{table}
\caption{Adverbial sense-class distribution across different novels}
\label{tab:table8}
\begin{center}
\begin{tabular}{rrr}

\bf Verb& \bf Adverb sense-class for Author-1 & \bf Adverb sense-class for Author-2\\
\hline
letn\={a} (To take rest )    &  Temporal   & Temporal, Measure   \\
kheln\={a} (To play) & Temporal    & Temporal, Measure, Force     \\
likhn\={a} (To write)& Measure   &  Temporal   \\
girn\={a} (To fall)  &  Temporal, Spatial, Force, Measure & Temporal, Force  \\
j\={a}n\={a} (To go) & Temporal  & Temporal, Measure\\
dena\={a} (To give) & Temporal, Measure &  Temporal\\
sunan\={a} (To listen)  & Measure  & Temporal, Measure\\
\hline
\end{tabular}
\end{center}
\end{table}
\section{Conclusion and Future Work}
In this paper we used Formal Ontology of Language to develop ontological resource for Hindi 
and Telugu. Logical forms of intensional senses were identified as type and class.
The validation of this resource was done using Cohen's Kappa that showed higher agreement. 
The resource was used for extracting adverbial class distribution of verbs, k{\=a}raka-verb sense-type distribution from corpus. 
We compared different corpora based on sense-type distribution of verbs. Novels of different authors were compared using sense-class
of adverbs.
The usage of different sense-class of adverbs across
different authors indicates different sociolingustics aspect.  However, this just covers a portion of a language. Adjectival and
k{\=a}raka points will give a deeper insight. 
Further, validation and enrichment of the resource is in progress.
Major work ahead is to find etymological, morphological and syntactic points.
The resource can be utilized for word sense disambiguation, synonimity measure, cultural studies. 
\bibliographystyle{splncs}
\bibliography{paper}
\end{document}